# Revealing the impact of social circumstances on the selection of cancer therapy through natural language processing of social work notes

Shenghuan Sun[1],
Travis Zack[1,4],
Christopher Y.K. Williams[1],
Atul J. Butte [1, 2, 3]
Madhumita Sushil[1*]

1. Bakar Computational Health Sciences Institute, University of California, San Francisco, San Francisco, CA, USA
2. Center for Data-driven Insights and Innovation, University of California, Office of the President, Oakland, CA, USA
3. Department of Pediatrics, University of California, San Francisco, CA, 94158, USA
4. Division of Hematology/Oncology, Department of Medicine, University of California, San Francisco, San Francisco, California, USA.

*Author to whom correspondence should be addressed.

Abstract word count: 254
Main contain word count: 3722

# ABSTRACT

### OBJECTIVE

We aimed to investigate the impact of social circumstances on cancer therapy selection using natural language processing to derive insights from social worker documentation.

### MATERIALS AND METHODS

We developed and employed a Bidirectional Encoder Representations from Transformers (BERT) based approach, using a hierarchical multi-step BERT model (BERT-MS) to predict the prescription of targeted cancer therapy to patients based solely on documentation by clinical social workers. Our corpus included free-text clinical social work notes, combined with medication prescription information, for all patients treated for breast cancer at UCSF between 2012 and 2021. We conducted a feature importance analysis to pinpoint the specific social circumstances that impact cancer therapy selection.

### RESULTS

Using only social work notes, we consistently predicted the administration of targeted therapies, suggesting systematic differences in treatment selection exist due to non-clinical factors. The UCSF-BERT model, pretrained on clinical text at UCSF, outperformed other publicly available language models with an AUROC of 0.675 and a Macro F1 score of 0.599. The UCSF BERT-MS model, capable of leveraging multiple pieces of notes, surpassed the UCSF-BERT model in both AUROC and Macro-F1. Our feature importance analysis identified several clinically intuitive social determinants of health (SDOH) that potentially contribute to disparities in treatment.

### DISCUSSION
Leveraging social work notes can be instrumental in identifying disparities in clinical decision-making. Hypotheses generated in an automated way could be used to guide patient-specific quality improvement interventions. Further validation with diverse clinical outcomes and prospective studies is essential.

### CONCLUSIONS
Our findings indicate that significant disparities exist among breast cancer patients receiving different types of therapies based on social determinants of health. Social work reports play a crucial role in understanding these disparities in clinical decision-making.

**OBJECTIVE**

Clinical decisions biased by social disparities lead to significant discrepancies in outcome and pose significant public health concerns [1–3]. Clinical decisions are influenced not only by clinical criteria but also by non-clinical factors such as race, gender, perceived financial stability, and more, which are collectively referred to as social determinants of health (SDOH)[4–6]. There is growing evidence that many minority groups are less likely to receive standard of care[6,9,10]. One pressing example is the decision to initiate anti-neoplastic treatments, which are becoming increasingly expensive and associated with financial toxicities[7]. While new, targeted agents often are better tolerated and more effective than previous treatments, they can come with a high price tag not always fully covered by insurance, leaving clinicians with a moral decision when balancing efficacy and cost. Financial constraints are but one example of factors that can potentially influence the treatment decision[8].

In this work, we demonstrated a strong association between specific features within social work (SW) clinical documentation and the choice of expensive, targeted therapy prescription for patients with breast cancer. Using a pretrained Bidirectional Encoder Representations from Transformers (BERT) model, we showed that the unstructured SW notes, without detailed diagnostic or therapeutic, can predict whether targeted therapy was prescribed for a given patient. Moreover, we developed a hierarchical language model for prediction over long sequences of clinical notes and successfully increased the predictability of the outcome. To understand which SDOH factors are used by the model for prediction, we measured the importance of SDOH factors by deleting words belonging to specific SDOH topics. Several critical contributors emerged, including socio-economic factors, abuse history, and risk of death. Our findings demonstrate that SW notes can reveal the impact of a patient's social environment on medical treatment prescription without requiring expensive and time-consuming manual annotation. Our hierarchical modeling approach will inform the development of models capable of leveraging multiple clinical notes for prediction.

**BACKGROUND AND SIGNIFICANCE**

A growing body of evidence indicates that SDOH factors significantly impact patient health and behaviors[5,6,11,12]. However, SDOH factors not only affect patients but also influence the clinical decision-making process recommended by physicians[4]. Ideally, clinical decision-making should be rooted in evidence-based practices, cognizant of the complex interaction between a patient's background and SDOH that could affect both their trust in the medical system and their overall disease trajectory. In reality, though, physicians are inevitably influenced by a wide range of non-clinical factors, with many of these non-clinical factors rooted in unconscious bias[13,14]. Previous research showed that clinical management decisions can be influenced by socioeconomic status[8], race[15], gender[16], adherence to treatment[17], patient behavior[18], attitude[19], and even physician personal characteristics[20].

Although it is well-known that SDOH-related, non-medical factors are crucial contributors to health and clinical outcomes, extracting non-medical and social factors from electronic medical records remains challenging. While information such as smoking, alcohol, and primary insurance status is increasingly accessible in structured fields, many social factors that are increasingly recognized as being important to successful treatment, are either not captured or are not a focus of structured physician documentation. Various aspects that physicians consider, including patient personalities, preferences, faith, concerns, professional interactions, family support, and living situations, can often be missing or improperly addressed within physician notes [4]. Due to this, our capacity to understand the relationship between these critical aspects of SDOH is constrained by the data we choose to focus on, as well as the accessibility of the information within.

Compared to general clinical documentation, notes written by social workers (SW notes) contain comprehensive social information[21,22]. Social workers are professionals who specialize in navigating a patient through the barriers that may interfere with receiving adequate medical care[23,24]. They can evaluate the many aspects of patients' life outside of medicine that can impact their ability to receive treatment. These include insurance concerns, financial concerns, social and daily living support, and ancillary support such as transportation, mental health, and housing. Because of this focus on the non-medical barriers that may affect medical care, SW notes could be invaluable in understanding the non-medical factors that influence medical decision-making.

Demonstrating that social work notes, considered in isolation, can be predictive of complex clinical decisions would highlight the power that can be derived from understanding how SDOH affects clinical decision-making. Doing this requires the development of new methods in natural language processing (NLP) to transform the nuances within SW documentation of complex social topics into predictive features around the clinical decision-making for costly drugs.

**MATERIALS AND METHODS**

**Study design and cohort selection**

This study used a deidentified clinical note corpus at UCSF available within the UCSF Information Commons. The research was conducted under the IRB #18-25163). Our corpus included the deidentified social work notes of all patients treated at UCSF for breast cancer between 2012 and 2021 (**Figure 1**). Breast cancer diagnosis was identified using the ICD9 code 174 and the ICD10 code C50 through the UCSF Clinical Data Warehouse. We obtained 2496 patients matching these codes, with available social work reports. We then retrieved the medications ordered or prescribed for these patients, then categorized these as "targeted therapy" medications or not based on the definitions in the *Targeted Cancer Therapies Fact Sheet from National Cancer Institute*[25]. Patients in the cohort received targeted therapy at least once were categorized as '*Targeted therapy administered*' group (TT-Yes); patients who did not receive any targeted therapy were categorized as '*Targeted therapy not administered*' group (TT-No). Our study design enabled us to investigate the impact of social circumstances on cancer therapy selection using natural language processing.

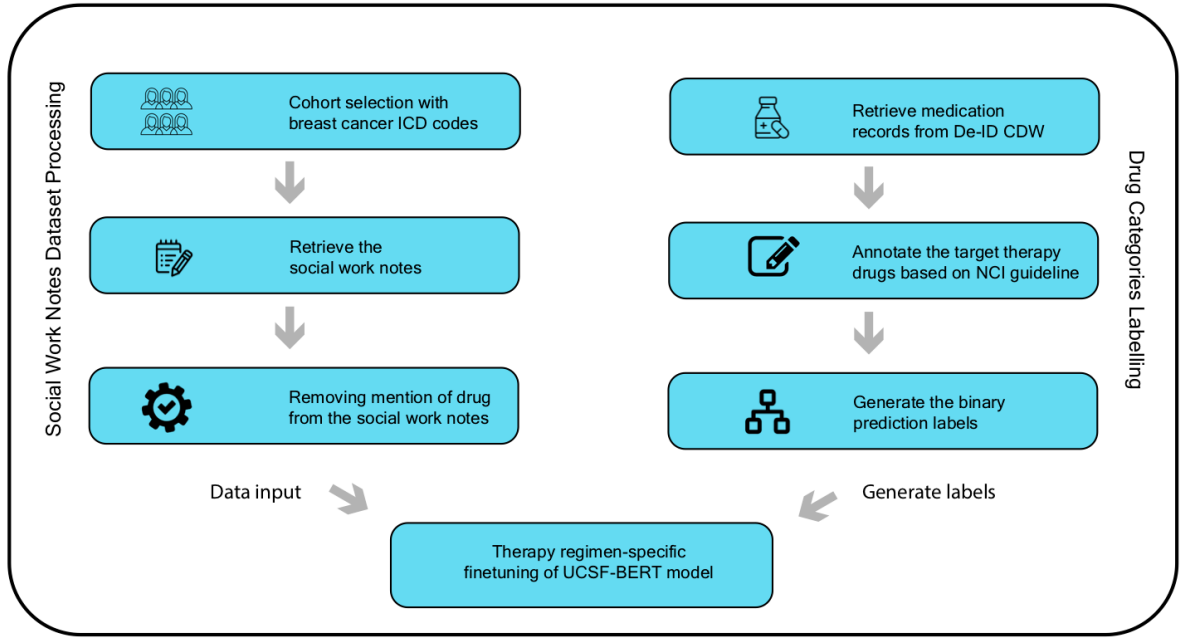

**Figure 1.** The overall workflow. We implement an end-to-end BERT-base classification model to predict the category of treatment administration for breast cancer patients at UCSF. We first retrieved the patients' social work notes from UCSF de-identified Caboodle Data Warehouse (DeID-CDW) between 2012 and 2021. We then annotated whether an individual patient has ever received targeted therapy based on the Targeted Cancer Therapies Fact Sheet from National Cancer Institute. In this manner, we obtained 2496 patients, of which 70% received targeted therapy. The dataset was further split into 8:1:1 ratios, corresponding to training, validation, and test sets.

**Deep learning models for sentence classification**

We used latest and longest social work note per patient to predict cancer therapy selection. Patients notes in our dataset were randomly split in an 8:1:1 ratio into training, validation and test sets. We trained our algorithm on the training set, using the early stopping approach to help with parameter tuning on the validation set. We ran our algorithm 5 times for each model and evaluated the model performance using the validation set. We reported the median scores in the five models on the test set to reduce the training variation. The cross-entropy loss function was used for optimization. After training and hyperparameter tuning, the model was tested on the held-out test set to compute model performance.

We used an internal UCSF-BERT model, which is a cased BERT model pretrained on the UCSF clinical notes using the masked language modeling (MLM) and next sentence prediction (NSP) objectives from scratch, developed by *Sushil et al* [26]. We selected this model with the hypothesis that pretraining on the same domain will improve the model's performance. We also implemented publicly available pretrained BERT-base models: SciBERT[28], ClinicalBERT[29], BioLM[30], and Biomed-Roberta[35]. We fine-tuned these models for the classification task.

To rule out the possibility of finding results by randomness, we implemented three distinct dummy classifiers as a control. Dummy (Prior): This strategy always predicts the most frequent class in the training set. Dummy (Stratified): This strategy generates predictions by respecting the class distribution of the training set. It randomly predicts class labels based on the distribution of the training set. Dummy (Uniform): This strategy generates predictions uniformly at random.

### Evaluation metrics

Model evaluation results were reported for the testing dataset only. For the classification task, Area Under the Receiver Operating Characteristic curve (AUROC), F1 score, precision, and recall metrics are reported. In order to address the issue of data imbalance, which can impede the interpretation of model performance, we used macro-averaged format for F1, precision, and recall score. F1 score is the harmonic mean of precision and recall.

Notably, macro-averaged computation uses the arithmetic mean of all the per-class scores, which provides equal weight to all the classes. We used *sklearn.metrics* from the scikit-learn python package for programming[32].

### Constructing the BERT-MS model

Although most patients in our dataset have several relevant SW notes (median = 11, **Figure 2B**), the BERT model used for classification is unable to accept more than a maximum of 512 tokens, which cannot handle more than one social work note piece. We were interested in knowing whether integrating more notes and thus more information about a patient's social history would improve the prediction. However, retraining a language model with an input length several times longer would take considerable time and computation resources and is impractical in an academic environment[27]. Consequently, we developed a multistep, hierarchical BERT model that can integrate several notes named UCSF BERT MS-n, where *n* refers to the maximum number of notes allowed by the model (**Figure 3**).

The UCSF BERT-MS-n model was trained in two steps. First, we used multiple notes for a single patient as multiple independent instances for classification. We then trained the UCSF-BERT classification model on this set. Second, we retained only the lower-dimensional hidden representations of each note from the final layer of the trained classification model obtained in the previous step. These low-dimensional representations of multiple reports of the same patient were thereby concatenated into a single representation. We then trained a second multilayer perceptron for the same classification task **(Figure 3)**. In this manner, we were able to train a hierarchical language model that can integrate n-fold information without expending the model parameter n-folds. We built several UCSF BERT-MS-n models including UCSF BERT-MS-3, UCSF BERT-MS-5, UCSF BERT-MS-8, UCSF BERT-MS-10, that corresponds to the use of at most 3, 5, 8 and 10 notes.

**Feature importance analysis**

To understand which SDOH factors are used by the model for prediction, we used feature ablation methods to measure the importance of different SDOH factors. We examined the effect on model performance of removing keywords associated with the following topics: Mental health, Family, Consultation/Appointment, Group session, Risk of death, Clinician/Hospital/Medication, Living condition/Lifestyle/Social support, Telephone encounter/Online communication, Abuse history (all forms), and Insurance/Income. These categories, and keywords associated with each category, were selected following the LDA topic modelling analysis as described by Sun et al[33] (Supplementary Table 5). Specifically, we removed a set of words belonging to each SDOH topic iteratively from the test set only and compared the decrease in model performance represented by the decrease in F1 score. We conducted these experiments on UCSF BERT MS-5 model which has the best predictive performance.

To account for differences in the prevalence of various topics mentioned across patients (e.g 96% of notes contained keywords in the 'Social support' topic whereas only 10% of notes relate to the 'Risk of death' category), we normalize the importance of each topic by their frequency. We present both the raw feature important score and the important score normalized by topic frequency in **Supplementary Figure 2.**

**RESULTS**

**Patients structured characteristics and their social work notes.**

We identified 2496 patients with breast cancer with available deidentified social work notes (**Figure 1**) with a distribution of 97.9% female and 2.1% male. There were 59.7% White/Caucasian patients, 18.1% Asian, 10.1% Hispanic/Latino, 6.5% Black/African, and 15.7% Other (**S. Table 1**). Overall, 70% of patients in *the* cohort received targeted therapy at least once [*'Targeted therapy administered'* group](TT-Yes), compared to 30% of patients who did not receive any targeted therapy [*'Targeted therapy not administered'* group](TT-No).

First, we explored whether SDOH information within structured data alone could stratify these patients. For the 2496 patients identified, we found information regarding demographics, marriage status and smoking history was present, but data on patient financial status, education level and other important SDOH were absent from the structured data. Machine learning-based approaches leveraging the SDOH coded within structured data failed to predict the administration of targeted therapy in patients (**S. Table 3**), demonstrating the limitations of structured data in predicting this task.

In contrast, our prior research has demonstrated that social work notes possess a wealth of information relating to SDOH, including details on frequently discussed topics such as mental health, insurance status and family support (**Figure 2C, D**)[33]. This qualitative observation suggested that social work notes encompass a wealth of SDOH factors, which may be captured by pre-trained language models when predicting the administration of therapy regimens to patients.

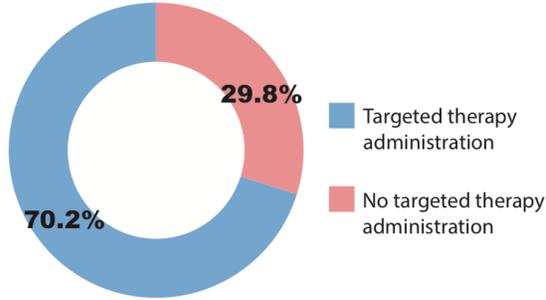
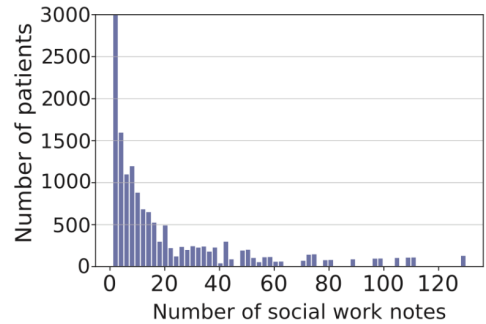

**Figure 2.** Data exploration on social work notes. A. Pie chart showing the different proportions of patients in the two categories. B. Histogram showing the number of notes for the individual patients (mode = 2, mean = 22, median = 11). C. Example deidentified social work notes. Top: Example patient who did not have any targeted therapy administration. Bottom: Example patient who received at least one dose of targeted therapy.

**UCSF-BERT outperforms public language models in predicting therapy regimen**

We fine-tuned the pretrained UCSF-BERT classification model[26] to predict the targeted therapy administration directly from the social work notes of breast cancer patients. The UCSF-BERT model incorporates notes from the same electronic health record from which our SW notes are derived and hence is best suited for our task. Given that the maximum sequence length supported by a regular BERT model is 512 tokens[27], we used the longest note for each patient to maximize the amount of information available for classification. For comparison, we implemented several publicly-available transformer-based language models[27–31]. They are all customized language models that have been trained on a large corpus of scientific texts, PubMed, PMC, or clinical notes from the MIMIC-III corpus[36]. **Table 1** shows the prediction performance of different deep-learning classification models. UCSF-BERT achieved the best result with a Macro F1 of 0.599 and AUROC score of 0.675. RoBERTa models (BioLM and Biomed-Roberta) performed generally better than BERT-base models (SciBERT, ClinicalBERT) potentially because of their dynamic masking strategy during pretraining such that the masked token changes during each training epoch[31]. Yet, none of them outperformed the UCSF-BERT model. This suggests that pretraining BERT-based models with clinical data can be helpful for achieving superior performance on domain-specific tasks. We also ran our tasks on three random baseline models, each of which ruled out the random performance from different perspectives **(See methods)**. Our model significantly outperformed the random baselines (Table 1). Overall, the predictive performance of the UCSF-BERT model suggested that the social circumstance of patients is associated with the type of therapies administered to them.

**Integrating multiple clinical notes for prediction**

Given that the median number of clinical social notes per patient in our cohort is 11, we built several UCSF BERT_MS-n models including UCSF BERT_MS-3, UCSF BERT_MS-5, UCSF BERT_MS-8, UCSF BERT_MS-10, allowing the analysis of up to 3, 5, 8, and 10 notes respectively. Table 2 compares the prediction performance of UCSF BERT_MS-n models with the original UCSF BERT model, which previously exhibited superior performance than public language models. Generally, the UCSF BERT_MS-n models achieved better results, demonstrating the advantage of incorporating more clinical notes.

**Table 1. Model performance of different classifiers.** UCSF-BERT achieved supervisor performance in AUC, MACRO F1, as well as MACRO RECALL.

| Model | AUC | MACRO F1 | MACRO PRECISION | MACRO RECALL |
|---|---|---|---|---|
| UCSF BERT | **0.675** | **0.599** | 0.604 | **0.596** |
| ClinicalBERT | 0.627 | 0.578 | 0.584 | 0.576 |
| SciBERT | 0.616 | 0.532 | 0.606 | 0.533 |
| BioLM | 0.671 | 0.583 | **0.615** | 0.580 |
| Biomed-RoBERTa | 0.667 | 0.584 | 0.592 | 0.581 |
| Dummy (Prior) | 0.500 | 0.412 | 0.350 | 0.491 |
| Dummy (stratified) | 0.504 | 0.525 | 0.529 | 0.603 |
| Dummy (Uniform) | 0.500 | 0.509 | 0.522 | 0.602 |

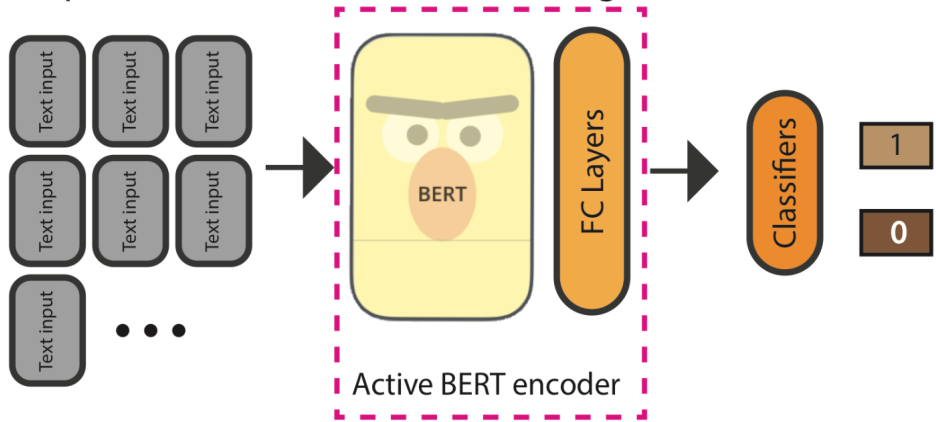

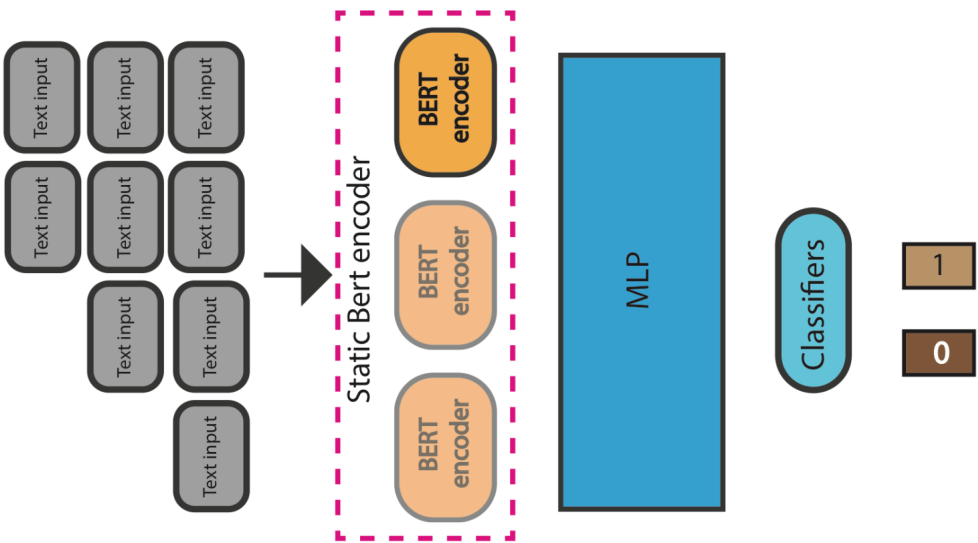

**Figure 3**: Illustration of BERT-MS-n model. To use long sequences of clinical notes for prediction, we built a hierarchical BERT model (BERT-MS), where the first step divides a long sequence of notes into multiple independent instances and then trains the single BERT classifier on the individual chunks in the training set. In the second step, we concatenate the BERT representations of all notes of the same patient and further fit them into a multilayer perceptron for the training. FC: Fully connected Layer; MLP: Multi-layer perceptron.

Table 2. **BERT MS model achieved superior performance** in AUC, MACRO F1, as well as MACRO RECALL.

|  | AUC | MACRO F1 | MACRO PRECISION | MACRO RECALL |
|---|---|---|---|---|
| UCSF BERT | 0.675 | 0.599 | 0.604 | 0.596 |
| UCSF BERT MS-3 | 0.707 | 0.620 | 0.660 | 0.612 |
| UCSF BERT MS-5 | 0.702 | **0.624** | 0.637 | 0.615 |
| UCSF BERT MS-8 | **0.718** | 0.623 | 0.645 | **0.616** |
| UCSF BERT MS-10 | 0.706 | 0.596 | **0.665** | 0.594 |

**Identifying the SDOH factors that influence model decisions.**

To explore the role different SDOH factors may have in predicting utilization of targeted therapy, we assessed the importance of SDOH factors by feature ablation methods (**See Methods**). The 11 topics that we tested were mentioned with varying frequency in the social work notes (**Supplementary Figure 1**). Notably, our findings did not reveal any topic specific class imbalance among the extracted topics in patients who were not administered targeted therapy. The notes belong in each topics have the similar class proportion: 70% "TT-Yes" group and 30% "TT-No" group.

We found several SDOH topics, including *Abuse History*, *Risk of Death*, *Social Support* as the most significant influencers that the model leveraged in the prediction task (**Figure 4**). Other SDOH topics such as *Family*, *Living Condition* also had obvious impact in model decision making. However, besides the broad topic area "medical factors", the common topics related to medical aspects, *Mental Health* and *Group Session*, had a lower influence on the model prediction. As the neutral control, topic *Consultation* and *TelephoneEncounter* played a less important role in the prediction task. Interestingly, Finance, which represents the socioeconomic factor that likely influences patients' decisions in therapy regimen, did not come up as an important regulator in the process. Overall, we successfully used model interpretability methods to analyze the trained language model to discover the SDOH factors that are not frequently considered to be influencers of the prescription of more financially toxic oncology medications.

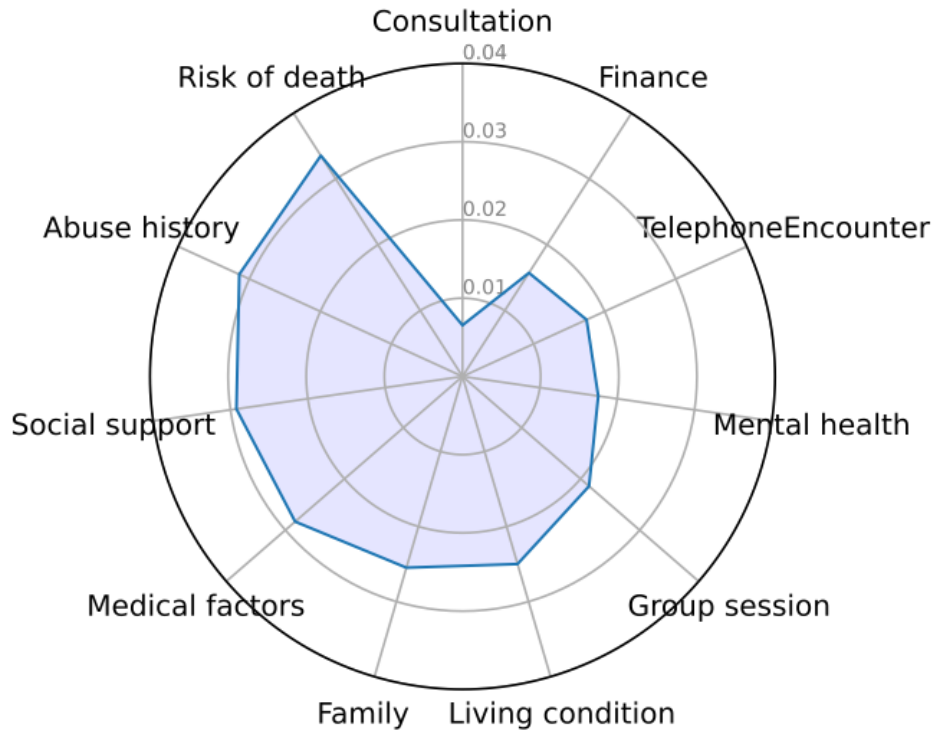

**Figure 4**. Feature importance analysis for SDOH factors in ablation study. The radar chart shows the feature importance of SDOH topics, representing by the decrease of F1 score.

**DISCUSSION**

This study demonstrated that clinical social work documentation, which focus more on social determinants of health rather than treatment plans, can be predictive of whether targeted therapies are administered to patients with breast cancer and highlights a potential SDOH-dependent disparity in therapy administration. Additionally, we developed a hierarchical modeling technique to incorporate the large volume of note data within any given chart, which often exceeds the processing capacity of the state-of-the-art NLP models. This technique can leverage multiple notes for prediction without adding a significant amount of computation burden. Finally, we performed a feature importance analysis by ablation of SDOH-related keywords to better understand which topics within social work notes have the greatest contribution to model performance.

We found that pretraining a language model on the same data source is important for better prediction performance in specialized domains, particularly from small datasets that are common in clinical studies. Among all the transformer-based models we explored (**Table 1**), UCSF-BERT achieved the best prediction performance on our task. Moreover, with our hierarchical BERT model, we showed that integrating multiple notes, and consequently more information about a patient, improves model performance. It is generally accepted that including more comprehensive patient information, either from clinical notes written at different times during a health encounter or for a different purpose, will lead to better performance for prediction tasks. Although alternate methods that allow longer input text exist, such as the Longfomer technique[38], these approaches usually require retraining a large language model, which can be time-consuming and computationally expensive.

In our feature importance analysis, we found that financial factors are not the sole SDOH factors influencing therapy regimen decisions, as initially hypothesized. Our feature importance analysis revealed other significant factors, including "risk of death" and "abuse history", led to decreases in model performance when removed from social work note text. The model's ability to consider the context of these mentions within sentences further emphasizes their importance. This study broadens our understanding of the various factors affecting therapy regimen choices, suggesting that a more comprehensive approach is needed when considering SDOH factors in clinical informatics. Future research should explore additional factors and their potential impact on therapy decisions to ensure a more holistic understanding of patient care.

There are several limitations to this study. While our research showed that social work reports that encompass SDOH information are predictive of the administered breast cancer therapy regimen, integration of structured data and other types of text reports may both highlight other aspects driving the disparity in treatment choice and improve overall predictive performance. While the aim of our paper is to demonstrate the utility of social work notes, comprehensively predicting therapy regimen decisions is complex and beyond the scope of our current paper. Systematically extracting and converting all SDOH factors from clinical notes to structured data may create additional opportunities for further analysis. In addition, our data was limited to cancer therapy treatment decisions at a single academic medical center. The driving forces behind treatment decisions for patients at other centers may be differ, as may the overall distribution of SDOH factors themselves. Patients may already be preselected in unrecorded ways to be seen by a social worker. Future work should seek to integrate data across institutions with differing practices to further validate our findings. Finally, while our methods for model interpretability are able to uncover important social topics that are associated with the observed disparity, the methods for post-hoc interpretability may not be entirely faithful to the originally trained model, which is an inherent limitation of the current state-of-the-art methods in NLP[39].

**CONCLUSION**

In conclusion, our study demonstrates the potential of utilizing transformer-based deep learning approaches for predicting clinical outcomes using social work reports. Specifically, our findings indicate the presence of notable disparities in treatment regimens, which can be attributed to social determinants of health. By creating a hierarchical model that can incorporate additional notes, we observed an enhancement in overall model performance. Through the use of ablation methods to better understand model interpretability, we highlighted the variety of SDOH factors that can influence therapy regimen selection for patients with breast cancer. Future research should extend this analysis to explore the impact of SDOH on treatment selection at other institutions and for different types of cancer.

**AUTHOR CONTRIBUTION**

AJB put forward the research idea. SS, MS and AJB designed the study. SS developed the methods, analyzed the data, and drafted the manuscript. AJB and MS supervised the study. All authors contributed to manuscript review and editing.

**ACKNOWLEDGEMENTS**

We thank all researchers, clinicians, and social workers who help collect clinical notes data in our UCSF Information Commons. We thank everyone in Dr. Atul J. Butte's lab for helpful discussion and feedback. We thank staff members in the Bakar Computational Health Sciences Institute and UCSF IT Services who build and maintain the UCSF Information Commons. We thank the Wynton High-Performance Computing (HPC) cluster support team for making available the needed computation capacity.

**COMPETING INTERESTS**

AJB is a co-founder and consultant to Personalis and NuMedii; consultant to Mango Tree Corporation, and in the recent past, Samsung, 10x Genomics, Helix, Pathway Genomics, and Verinata (Illumina); has served on paid advisory panels or boards for Geisinger Health, Regenstrief Institute, Gerson Lehman Group, AlphaSights, Covance, Novartis, Genentech, and Merck, and Roche; is a shareholder in Personalis and NuMedii; is a minor shareholder in Apple, Meta (Facebook), Alphabet (Google), Microsoft, Amazon, Snap, 10x Genomics, Illumina, Regeneron, Sanofi, Pfizer, Royalty Pharma, Moderna, Sutro, Doximity, BioNtech, Invitae, Pacific Biosciences, Editas Medicine, Nuna Health, Assay Depot, and Vet24seven, and several other non-health related companies and mutual funds; and has received honoraria and travel reimbursement for invited talks from Johnson and Johnson, Roche, Genentech, Pfizer, Merck, Lilly, Takeda, Varian, Mars, Siemens, Optum, Abbott, Celgene, AstraZeneca, AbbVie, Westat, and many academic institutions, medical or disease specific foundations and associations, and health systems. AJB receives royalty payments through Stanford University, for several patents and other disclosures licensed to NuMedii and Personalis. AJB's research has been funded by NIH, Peraton (as the prime on an NIH contract), Genentech, Johnson and Johnson, FDA, Robert Wood Johnson Foundation, Leon Lowenstein Foundation, Intervalien Foundation, Priscilla Chan and Mark Zuckerberg, the Barbara and Gerson Bakar Foundation, and in the recent past, the March of Dimes, Juvenile Diabetes Research Foundation, California Governor's Office of Planning and Research, California Institute for Regenerative Medicine, L'Oreal, and Progenity. None of these entities had any bearing on the design of this study or the writing of the manuscript.

**DATA AVAILBILTY**

The data that support the findings of this study are available from the Information Commons platform at UCSF, but restrictions apply to the availability of these data, which were used under license for the current study, and so are not publicly available. Data are however available from the authors upon reasonable request and with permission of UCSF.

## Supplementary data

S. Table 1: Demographic characteristics for breast cancer patients in our cohort.

|  | Targeted therapy not administered (TT-No, N=597) | Targeted therapy administered (TT-Yes, N=1899) | Overall (N=2496) |
|---|---|---|---|
| **Sex** | | | |
| Female | 572 (95.8%) | 1871 (98.5%) | 2443 (97.9%) |
| Male | 25 (4.2%) | 28 (1.5%) | 53 (2.1%) |
| **Ethnicity** | | | |
| Hispanic/Latino | 72 (12.1%) | 180 (9.5%) | 252 (10.1%) |
| Not Hispanic or Latino | 497 (83.2%) | 1646 (86.7%) | 2143 (85.9%) |
| Other | 28 (4.6%) | 73 (2.0%) | 101 (2.2%) |
| **Race** | | | |
| Asian | 110 (18.4%) | 341 (18.0%) | 451 (18.1%) |
| Black/African | 33 (5.5%) | 129 (6.8%) | 162 (6.5%) |
| White or Caucasian | 355 (59.5%) | 1136 (59.8%) | 1491 (59.7%) |
| Other | 99 (17.6%) | 283 (15.4%) | 382 (15.7%) |

**S. Table 2**: Summary characteristics of social factors (smoking and marital status) for breast cancer patients extracted from structured data

|  | Targeted therapy not administered (TT-No, N=597) | Targeted therapy administered (TT-Yes, N=1899) | Overall (N=2496) |
|---|---|---|---|
| **Smoking status** | | | |
| Current Every day Smoker | 11 (1.8%) | 39 (2.1%) | 50 (2.0%) |
| Current Some Day Smoker | 6 (1.0%) | 21 (1.1%) | 27 (1.1%) |
| Former Smoker | 177 (29.6%) | 569 (30.0%) | 746 (29.9%) |
| Never Assessed | 2 (0.3%) | 9 (0.5%) | 11 (0.4%) |
| Never Smoker | 395 (66.2%) | 1232 (64.9%) | 1627 (65.2%) |
| Passive Smoke Exposure - Never Smoker | 5 (0.8%) | 17 (0.9%) | 22 (0.9%) |
| Smoker, Current Status Unknown | 1 (0.2%) | 1 (0.1%) | 2 (0.1%) |
| *Unknown | 0 (0%) | 6 (0.3%) | 6 (0.2%) |
| Light Tobacco Smoker | 0 (0%) | 3 (0.2%) | 3 (0.1%) |
| Unknown If Ever Smoked | 0 (0%) | 2 (0.1%) | 2 (0.1%) |
| **Marital status** | | | |
| *Unspecified | 1 (0.2%) | 0 (0%) | 1 (0.0%) |
| Divorced | 65 (10.9%) | 212 (11.2%) | 277 (11.1%) |
| Legally Separated | 4 (0.7%) | 15 (0.8%) | 19 (0.8%) |
| Married | 286 (47.9%) | 909 (47.9%) | 1195 (47.9%) |
| Registered Domestic Partner | 7 (1.2%) | 9 (0.5%) | 16 (0.6%) |
| Significant Other | 14 (2.3%) | 37 (1.9%) | 51 (2.0%) |
| Single | 164 (27.5%) | 480 (25.3%) | 644 (25.8%) |
| Unknown/Declined | 18 (3.0%) | 43 (2.3%) | 61 (2.4%) |
| Widowed | 38 (6.4%) | 193 (10.2%) | 231 (9.3%) |
| RDP-Widow | 0 (0%) | 1 (0.1%) | 1 (0.0%) |

**S. Table 3.** Model performances of common machine learning classifiers using SDOH related structured tabular data on targeted therapy administration.

| | AUC | MACRO F1 | MACRO PRECISION | MACRO RECALL |
|---|---|---|---|---|
| KNeighborsClassifier | 0.497 | 0.491 | 0.496 | 0.497 |
| SVM Classifier | 0.500 | 0.434 | 0.383 | 0.500 |
| RandomForestClassifier | 0.519 | 0.483 | 0.592 | 0.517 |
| GradientBoostingClassifier | 0.509 | 0.458 | 0.635 | 0.509 |

**S. Table 4.** The properties of notes for breast cancer patient's cohort. (Measure the tokes length and compare with 512 tokens)

| Percentage of notes longer than >300 words | Percentage of notes longer than >400 words | Percentage of notes longer than >2000 characters | Percentage of notes longer than >2500 characters |
|---|---|---|---|
| 34.0% | 22.2% | 26.1% | 17.9% |

**S. Table 5:** The words in the *Keywords* column are the representative words used to define the topics

| Topics | Keywords |
|---|---|
| Mental health | mental, depression, anxiety, mood, psychological, physical, cognitive, emotional, mind, psychiatric |
| Family | family, parent, father, mother, child, children, sister, parents, relatives, clan, childhood, friends |
| Consultation/Appointment | appointment, consultation, consult, questionnaire, question, advice, biographical, wikipedia, relevant, questions, know, documentation |
| Group session | group, intervention, session, interpers, community, class, organization, together, part, organization |
| Risk of death | suicide, suicidal, risk, crisis, homicide, murder, commit, bombing, murdered, murders, bomber, killing, convicted, victims |
| Medical factors | patient, medication, hospital, medical, clinic, clinician, treatment, therapy, surgery, symptoms, patients, drugs, diagnosis, treatments, prescribed |
| Living condition/Lifestyle | shelter, housing, house, living, sleep, bedtime, building, buildings, urban, employment, suburban, campus, acres |
| Social support | social, service, support, referral, recommendation, recommend, worker, resource, supports, provide, supporting, supported, allow, providing, assistance, benefit, help |
| TelephoneEcounter/Online communication | telehealth, phone, call, video, telephone, mobile, wireless, gsm, cellular, dial, email, calling, networks, calls, messages, telephones, internet |
| Abuse history | abuse, history, addiction, alcohol, drugs, allegations, victim, violence, sexual, rape, dependence |
| Insurance/Income | insurance, income, coverage, financial, contracts, banking, finance, liability, private, pay |

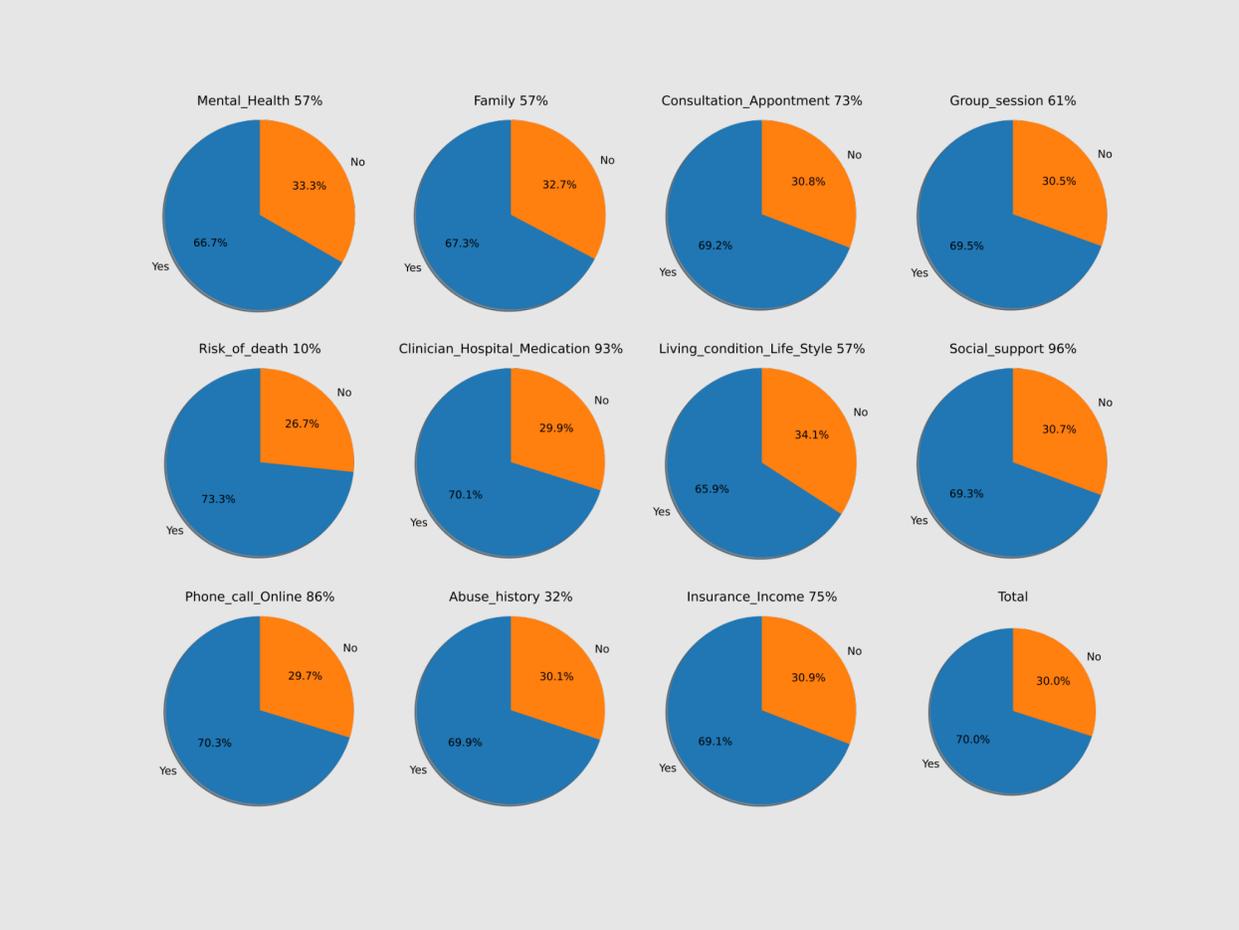

**Supplementary Figure 1.** Pie chart showing the different proportions of patients in the two categories per topic category. The percentage on the right of each topic indicates the frequency of whether words in the topics existed in individual social work notes. Orange: Patients who did not receive any targeted therapy. Blue: Patients who received at least one dose of targeted therapy.

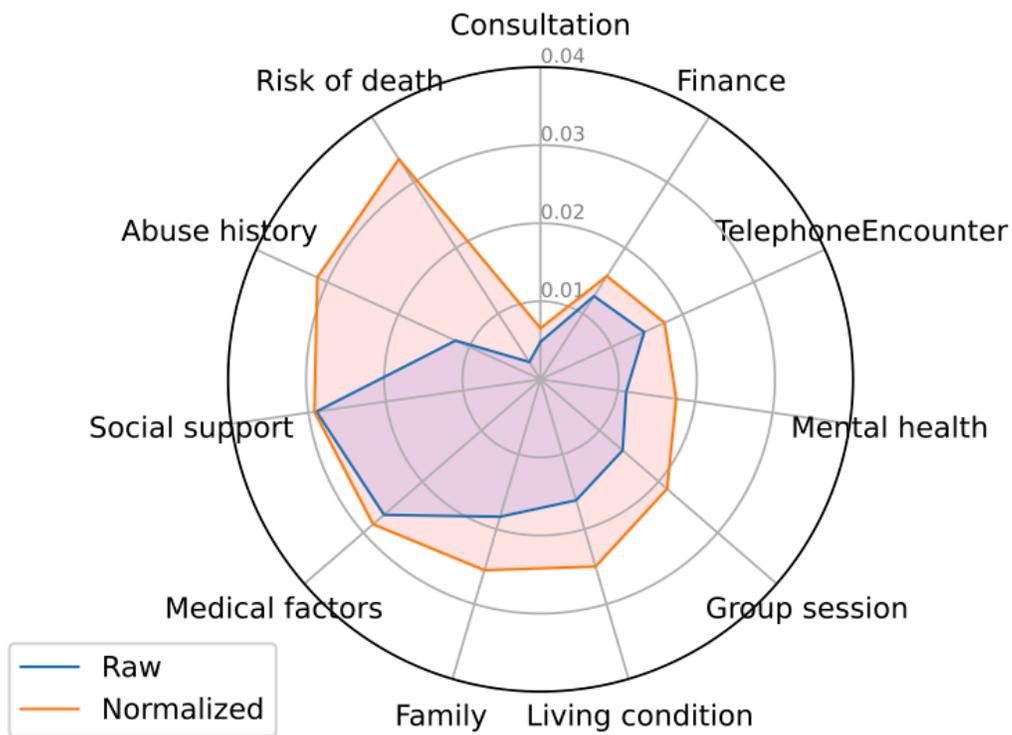

**Supplementary Figure 2**. Feature importance analysis for SDOH factors. The radar chart shows the feature importance of SDOH topics. Feature importance is defined to be the decrease in the F1 score of *Targeted therapy not administered* class across the entire test set overall (Raw: Blue), and across the notes that actually contain these words (Normalized: Orange).

'Social Work Note 08/22/2015 PT previously seen by {…} Family composition/living situation: PT is originally from China, but has lived in the ***** ***** for over ten years. Pt lives w/ her in-***** in ***** and husband and 6yo daughter. Legal: No legal issues identified. Emotional status/coping: PT endorsed stable mood during ***** *****. Support: PT describes FOB as supportive, though they do not have much time together d/t his work schedule Risk Factors: hx of depression; PT describes in-***** as verbally abusive (denies physical abuse); Limited social support Finances/Employment: PT is receptionist at a dental clinic. {…}"'

'Data: Ms. ***** was referred to ***** ***** ***** Work Services for assistance with resources related to Domestic Violence. (See note from 05/11/15) Her husband pushed her last week and also pushed her in October. She states that she has not "hit" her (hx of pushing 3x) SW provided NP referral info last week during the medical appointment.

'Social Work Data: SW referred to this pt by team pharmacist {…} Pt confirmed that she used the words "mean" and "controlling" to describe her husband when she was speaking to the team pharmacist. She elaborated to provide me with numerous examples of instances when she felt that her husband was "*****" and "putting [me] down". Pt said her husband has been like this for the entirety of their marriage and that he has current health issues ("memory impairment" and "is deaf") that have caused him to be more abusive as of late. {…}"'

**Supplementary Figure 3**. Example deidentified social work notes contain abusive history information.